# QuadSim: A Quadcopter Rotational Dynamics Simulation Framework For Reinforcement Learning Algorithms

Burak Han Demirbilek
Aselsan Research Center
Aselsan Inc.
Ankara, Turkey
bhdemirbilek@aselsan.com.tr

*Abstract*—This study focuses on designing and developing a mathematically based quadcopter rotational dynamics simulation framework for testing reinforcement learning (RL) algorithms in many flexible configurations. The design of the simulation framework aims to simulate both linear and nonlinear representations of a quadcopter by solving initial value problems for ordinary differential equation (ODE) systems. In addition, the simulation environment is capable of making the simulation deterministic/stochastic by adding random Gaussian noise in the forms of process and measurement noises. In order to ensure that the scope of this simulation environment is not limited only with our own RL algorithms, the simulation environment has been expanded to be compatible with the OpenAI Gym toolkit. The framework also supports multiprocessing capabilities to run simulation environments simultaneously in parallel. To test these capabilities, many state-of-the-art deep RL algorithms were trained in this simulation framework and the results were compared in detail.

## I. INTRODUCTION

Deep reinforcement learning is a subfield of the domains of machine learning, optimal control and decision-making. With recent years, the theory and the applicability have been extended to use in many research areas, including intelligent control. Research and development on real-life applications needs resources such as time or money, and therefore the use of simulations is necessary to avoid such costs. In this study, the reference tracking control task of a quadcopter is investigated, and a simulation framework has been designed to allow researchers to use reinforcement learning algorithms (and as well other algorithms) in quadcopter based control research.

At the present time, there exists many quadcopter simulation frameworks such as [1], [2], [3] and they can provide many features and capabilities that are adequate to many research topics. For our research needs, we need a simple simulator where the users can start developing in minutes (aims to be plug-and-play) and every aspect of the source code can be customizable. The simulator is designed to test control algorithms for many configurations and dynamics using reinforcement learning algorithms in accordance with real-life pre-flight check procedures. The QuadSim simulation framework aims to support linear/nonlinear dynamics, deterministic/stochastic simulations, OpenAI Gym Toolkit [4] compatibility to allow many open-source algorithms to be run without any effort (such as Stable Baselines 3 [5] which has been used in the experiments section) and lastly, the single/parallel thread training sessions for high performance.

The layout of this paper is as follows: Firstly, in Section II, the problem will be defined and mathematically modeled. In Section III, the simulation framework will be explained with the design choices and experimental studies. After that the design of the performance metrics will be explained in Section IV and then the training framework, results and comparisons will be in the Section V. Lastly, the conclusion and future works will explain in the Section VI.

## II. PROBLEM DEFINITION AND MODELLING

The main focus of this simulation framework is to allow users to design and develop intelligent controllers within the scope of controlling angular rates. For this reason, only the rotational dynamics of the quadcopter are taken into consideration because of the indoor applicability purposes. The real-life applicability of this simulation framework is to test quadcopters on platforms (also known as testbeds) [6] with 3 DoF (degrees of freedom) and therefore, some assumptions/limitations has been made to both simulation and real-life studies applicable and easy.

For the mathematical definition and modelling, this work is verified with the similar studies [7] in the quadcopter dynamics literature to ensure that the mathematical modelling is correct. Below, you can see the assumptions made to make sure this mathematical model is correct:

- The rotational motion of the quadcopter is independent of its translational motion.
- The center of gravity coincides with the origin of the body-fixed frame.
- The structure of the quadcopter is rigid and symmetrical, with the four arms coinciding with the body x- and y-axes.
- Drag and thrust forces are proportional to the square of the propeller's speed.
- The propellers are rigid.

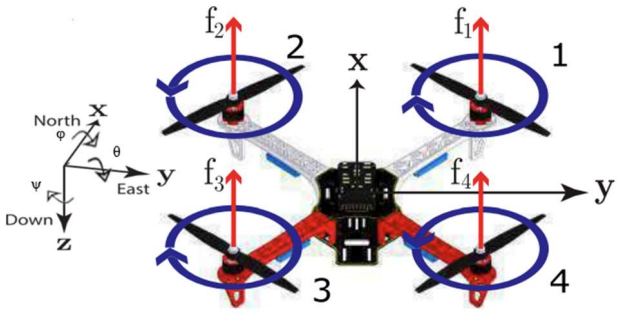

*Figure 1 - Earth-fixed and body-fixed coordinate frames, motor labels and quadcopter euler angles (Adapted from [19])*

The mathematical formulation starts with definition of some coordinate frames. Coordinate frames are needed to describe the motions of the quadcopter before the quadcopter is mathematically modeled.

### A. Coordinate Frames

We define two coordinate frames, the earth fixed frame (notated as $W$) and the body fixed frame (notated as $B$):

The earth-fixed frame is taken as the reference frame using the NED (North East Down) convention where the x-axis of the frame is pointed to the north. The orientation of the quadcopter, known as attitude, is expressed in the body-fixed frame by Euler angles phi, theta, psi which corresponds to the roll, pitch and yaw angles. In order to relate the orientation of the quadcopter in the earth-fixed frame, a rotation matrix is defined. Below, the rotation matrix R is defined, it converts from the body-fixed frame to the earth-fixed frame (The notation of $C_x$ and $S_x$ are the abbreviated versions of the $\cos(x)$ and $\sin(x)$ operators):

$$R = \begin{bmatrix} C\psi\, C\theta & C\psi\, S\theta\, S\varphi - S\psi\, C\varphi & C\psi\, S\theta\, C\varphi + S\psi\, S\varphi \\ S\psi\, C\theta & S\psi\, S\theta\, S\varphi + C\psi\, C\varphi & S\psi\, S\theta\, C\varphi - C\psi\, S\varphi \\ -S\theta & C\theta\, S\varphi & C\theta\, C\varphi \end{bmatrix}$$

With the rotation matrix, you can calculate the earth-fixed position of any given position.

$$v' = Rv \qquad v = \begin{bmatrix} x \\ y \\ z \end{bmatrix}$$

After that, the body frame can be defined. A typical quadcopter is driven by four rotors on the four arms of the frame. The position of the arms according to the body frame can differ. Generally, the most used configurations are "X" and "+" which represents the shape of x and plus shape.

In this study, the "X" shape frame was used and below, you can see the numerated rotors (from 1 to 4), rotation of the numerated rotors (clockwise or counter cw) and the body frame in Figure 1. The defined roll, pitch and yaw angles (notated as $\phi, \theta, \psi$) are also depicted in Figure 2Figure 1 as the rotational elements on the x-y-z coordinate axes respectively.

### B. Dynamics Formulation

From the rotational equations of motion, quadcopter dynamics can be formulated, for these dynamics, we firstly define the inputs as:

$$U_1 = db(w_4^2 - w_2^2)$$
$$U_2 = db(w_1^2 - w_3^2)$$
$$U_3 = k(w_1^2 + w_3^2 - w_2^2 - w_4^2)$$

Where:
- b - thrust coefficient
- k - aerodrag coefficient
- d - moment arm
- w - motor speed

After that, the nonlinear system dynamics can be written from the rotational equations of motion as:

$$\ddot{\phi} = \frac{(I_{yy} - I_{zz})\dot{\theta}\dot{\psi}}{I_{xx}} + \frac{U_1}{I_{xx}}$$

$$\ddot{\theta} = \frac{(I_{zz} - I_{xx})\dot{\phi}\dot{\psi}}{I_{yy}} + \frac{U_2}{I_{yy}}$$

$$\ddot{\psi} = \frac{(I_{xx} - I_{yy})\dot{\phi}\dot{\theta}}{I_{zz}} + \frac{U_3}{I_{zz}}$$

For both linear and nonlinear systems, below you can find the state and input definitions:

$$x = \begin{bmatrix} \varphi \\ \dot{\varphi} \\ \theta \\ \dot{\theta} \\ \psi \\ \dot{\psi} \end{bmatrix}, \dot{x} = \begin{bmatrix} \dot{\varphi} \\ \ddot{\varphi} \\ \dot{\theta} \\ \ddot{\theta} \\ \dot{\psi} \\ \ddot{\psi} \end{bmatrix}$$

$$u = \begin{bmatrix} U1 \\ U2 \\ U3 \end{bmatrix} = \begin{bmatrix} db(w_4^2 - w_2^2) \\ db(w_1^2 - w_3^2) \\ k(w_1^2 + w_3^2 - w_2^2 - w_4^2) \end{bmatrix}$$

### C. State Space (Linear) Representation Of The Quadcopter System

An LTI state space model is written below for model analysis and numerical calculations.

$$\dot{x} = Ax + Bu$$
$$y = Cx + Du$$

The system was linearized at the hover position, and it has been proven that the system is fully observable [7].

$$A = \begin{bmatrix} 0 & 1 & 0 & 0 & 0 & 0 \\ 0 & 0 & 0 & 0 & 0 & 0 \\ 0 & 0 & 0 & 1 & 0 & 0 \\ 0 & 0 & 0 & 0 & 0 & 0 \\ 0 & 0 & 0 & 0 & 0 & 1 \\ 0 & 0 & 0 & 0 & 0 & 0 \end{bmatrix}, B = \begin{bmatrix} 0 & 0 & 0 \\ \frac{1}{I_{xx}} & 0 & 0 \\ 0 & 0 & 0 \\ 0 & \frac{1}{I_{yy}} & 0 \\ 0 & 0 & 0 \\ 0 & 0 & \frac{1}{I_{zz}} \end{bmatrix}$$

$$C = \begin{bmatrix} 1 & 0 & 0 & 0 & 0 & 0 \\ 0 & 1 & 0 & 0 & 0 & 0 \\ 0 & 0 & 1 & 0 & 0 & 0 \\ 0 & 0 & 0 & 1 & 0 & 0 \\ 0 & 0 & 0 & 0 & 1 & 0 \\ 0 & 0 & 0 & 0 & 0 & 1 \end{bmatrix}, D = \begin{bmatrix} 0 & 0 & 0 \\ 0 & 0 & 0 \\ 0 & 0 & 0 \\ 0 & 0 & 0 \\ 0 & 0 & 0 \\ 0 & 0 & 0 \end{bmatrix}$$

Where the state vector x and the input vector $u$ is defined like this:

$$x = \begin{bmatrix} \phi \\ \dot{\phi} \\ \theta \\ \dot{\theta} \\ \psi \\ \dot{\psi} \end{bmatrix}, \dot{x} = \begin{bmatrix} \dot{\phi} \\ \ddot{\phi} \\ \dot{\theta} \\ \ddot{\theta} \\ \dot{\psi} \\ \ddot{\psi} \end{bmatrix}$$

$$u = \begin{bmatrix} U1 \\ U2 \\ U3 \end{bmatrix} = \begin{bmatrix} db(w_4^2 - w_2^2) \\ db(w_1^2 - w_3^2) \\ k(w_1^2 + w_3^2 - w_2^2 - w_4^2) \end{bmatrix}$$

### D. Nonlinear Dynamics

The summarized nonlinear dynamics can be written as:
$$\dot{x} = f(x, u)$$
$$y = g(x, u)$$

$$\dot{x} = f(x, u) = \begin{bmatrix} \dot{\phi} \\ \dfrac{(I_{yy} - I_{zz})\dot{\theta}\dot{\psi}}{I_{xx}} + \dfrac{U_1}{I_{xx}} \\ \dot{\theta} \\ \dfrac{(I_{zz} - I_{xx})\dot{\phi}\dot{\psi}}{I_{yy}} + \dfrac{U_2}{I_{yy}} \\ \dot{\psi} \\ \dfrac{(I_{xx} - I_{yy})\dot{\phi}\dot{\theta}}{I_{zz}} + \dfrac{U_3}{I_{zz}} \end{bmatrix}$$

$$y = g(x, u) = x = \begin{bmatrix} \phi \\ \dot{\phi} \\ \theta \\ \dot{\theta} \\ \psi \\ \dot{\psi} \end{bmatrix}$$

### III. SIMULATION FRAMEWORK

The framework definition is divided into 3 subsections and described in detail.

### A. Simulating/Solving The ODEs

To be able to compatible with the most recent machine learning tools, python programming language was selected to implement this simulation framework. Python is a high-level general-purpose programming language which can be used in many areas, and it is designed to be readable and easy to use. Since this design is limits the code execution performance when compared it with different programming languages, many extensions of Python have been released as modules (or packages) which can be installed on the python installation.

Scipy [8] is an open-source collection of optimized mathematical algorithms for python programming language. It includes algorithms from the topics of integration, optimization, interpolation, Fourier transforms, signal processing, linear algebra, etc. In this study, the Scipy ordinary differential equations solver functions were used with the "Explicit Runge-Kutta method of order 5" solver.

Within the solver parameters, this simulation aimed to have 250Hz simulation step frequency.

### B. OpenAI Gym Environment

To ease up the usage of the simulation for RL and other agents/algorithms, the simulation has been wrapped up with OpenAI Gym Toolkit [4]. Gym is a toolkit for comparing and developing reinforcement learning algorithms which is developed by OpenAI (open source). It can be used with the any numerical computation libraries such as Tensorflow [9], PyTorch [10] or Theano [11], and it contains collections of prebuilt test problems/environments which the researchers can use these environments or create their own by using the very basic class definitions. In summary, a gym environment consists of initialization, step and reset functions. Initialization function is responsible to create and initialize the environment with given parameters. A step function is the interaction function which iterates the simulation by 1 time step, returning observations/states with some other information. Reset function is to reset the simulation by setting internal variables and calling other functions, which it makes the simulation episodic.

By making simulations in accordance with these 3 basic functions, the generalized problem structure enables many algorithms to be applied into the structured simulation and with this way, the linkage between the simulation and the algorithms can be separated. For these reasons, OpenAI Gym toolkit was used in the simulation design to allow many algorithms can be run on the QuadSim environments (simulations).

#### 1) Environment States

With the given system dynamics, the aim is to track reference angle signals(roll, pitch and yaw angles) in the bestway. Therefore, we augment the state with extra 6-dimensional information which tells what reference signals the controller must follow.

To summarize, the augmented state is defined like this:

$$x = \begin{bmatrix} \phi_{err} \\ \dot{\phi}_{err} \\ \theta_{err} \\ \dot{\theta}_{err} \\ \psi_{err} \\ \dot{\psi}_{err} \end{bmatrix}$$

Note that error variables are defined as the difference between the reference state and current state. The angle states are also mapped between [-pi, pi).

#### 2) Environment Actions

The actions are defined in the same way as the original definition, actions are a vector with size (3,) and it tells the torque vector acting on the quadcopter in the body-fixed frame.

$$U = \begin{bmatrix} u_1 \\ u_2 \\ u_3 \end{bmatrix} = \begin{bmatrix} \tau_\phi \\ \tau_\theta \\ \tau_\psi \end{bmatrix}$$

The environment will automatically calculate the propeller speeds with its motor mixing algorithm. So, the agent only needs to give the environment the torque vector.

#### 3) Environment Limits (min and max)

After defining states and actions, the size of the state and action space needs to be defined. Since states and actions are not limited in theory, we can limit them by specify the duration of the simulation, defining $t_{start}, t_{end}$. This will bound the state and action space in terms of time duration, and limit values can be calculated when the time range is defined in the simulation initialization.

$$Thrust = b \sum_{i=1}^{4} w_i^2 = F = m * g$$

$$w_{min} = w_i = \sqrt{\frac{m*g}{4*b}}$$

$w_{max}$ = Can be taken from the dc motor datasheets.

To convert these units of RPM (rotation per minute) to radians per second, the simple conversion equation can be used

$$rad/s = \frac{RPM * 2\pi}{60}$$

From the motor min-max speed values, we can calculate the $U_{min}$ and $U_{max}$ values:

$$U_{max} = \begin{bmatrix} U_1 \\ U_2 \\ U_3 \end{bmatrix} = \begin{bmatrix} db((w_{max}^2 - w_{min}^2)) \\ db((w_{max}^2 - w_{min}^2)) \\ 2k((w_{max}^2 - w_{min}^2)) \end{bmatrix}$$

$$U_{min} = -U_{max}$$

After that, we can calculate maximum state values when the motor speeds are constant by calculating the definite integral over this period. Please note that rotational values are already bounded by pi, so we don't need to calculate the maximum value of them.

$$\varphi_{max} = \pi \; rad$$
$$\dot{\varphi}_{max} = \frac{U_{1max}}{I_{xx}} * (t_{end} - t_{start}) \; rad/s$$
$$\theta_{max} = \pi \; rad$$
$$\dot{\theta}_{max} = \frac{U_{2max}}{I_{yy}} * (t_{end} - t_{start}) \; rad/s$$
$$\psi_{max} = \pi \; rad$$
$$\dot{\psi}_{max} = \frac{U_{3max}}{I_{zz}} * (t_{end} - t_{start}) rad/s$$

As these hard limits could not be reached in any real-life application, since the calculation does not include any physical limit, only the linear model is considered, additional soft limits are defined to have more accurate limitations, these limitations are used from the Pixhawk 2.1 Cube IMU Sensor(MPU-9250) Data sheet [12], which supports measurements up to 2000 degree/second (~35 radian/second)

$$\varphi_{max} = \pi \; rad$$
$$\dot{\varphi}_{max} = 35 \; rad/s$$
$$\theta_{max} = \pi \; rad$$
$$\dot{\theta}_{max} = 35 \; rad/s$$
$$\psi_{max} = \pi \; rad$$
$$\dot{\psi}_{max} = 35 \; rad/s$$

*4) Simulation Frequency*

For this simulation, a simulation frequency of 250Hz and controller (control algorithm) frequency of 50Hz have been selected and used.

From the OpenAI Gym perspective, the action selected from the algorithm will be constant for every 0.02(1/50) seconds.

*5) Reward/Cost Function*

Since the aim of this control problem is tracking the reference signals, a reward/cost function is defined to be a quadratic cost of error, error means that the difference between the reference and current state. With such quadratic cost, the RL algorithms can be compared and criticized with the optimal control algorithms.

Below, this definition is expressed as a formal mathematical expression.

$$\bar{x} = x_{ref} - x_{current}$$
$$cost = \bar{x}^T Q \bar{x} + u^T R u$$
$$Q = identity(6,6) / \text{max state values(soft limits)}$$
$$R = identity(3,3) / \text{max action values(soft limits)}$$
$$reward = -cost$$

Where $Q$ and $R$ are reward/cost matrices, which are assumed to be normalized identity matrices of shape (6,6) and (3,3) respectively. The shape of the current and reference states vectors are (6,1). The term of soft limits means that soft limit values (which is explained in section 3) have been used in this normalization because of the theoretic hard limits are unrealistic and large since no physical assumptions (air friction, material properties, etc.) made about these values. In empirical experiments, it has been observed that the change in reward due to normalization is very small at low speeds, and this affects learning performance. In order to solve this problem, normalization is done by using the soft limits of the system.

*6) Reset Function*

In each reset, reference states of phi, theta and psi states are randomly generated in range [-pi, pi]. The current state is then calculated with the difference of reference state and current dynamics state. Also, the quadcopter dynamic states are not set to reset in each call to this function. The only case for internal dynamics to be reset is when the accumulated states exceed the soft limits of the environment.

$$x_{ref} = random(6,1)$$
$$x_{current} = \text{initial states if } x_{current} \text{ exceeds soft limits}$$
$$\bar{x} = x_{ref} - x_{current}$$

| Parameter | Default Value |
|---|---|
| Moment of inertia about x-axis, Ixx | $0.0213 \; kgm^2$ |
| Moment of inertia about y-axis, Iyy | $0.02217 \; kgm^2$ |
| Moment of inertia about z-axis, Izz | $0.0282 \; kgm^2$ |
| Mass, m | $1.587 \; kg$ |
| Gravity, g | $9.81 \; N$ |
| Moment arm, d | $0.243 \; m$ |
| Thrust coefficient, b | $3.7102e\text{-}5 \; Ns^2$ |
| Drag coefficient, k | $7.6933e\text{-}7 \; Nms^2$ |
| Propeller maximum angular speed, $w_{max}$ | $494.27 \; rad/s$ |
| Soft phidot limit, $\dot{\phi}$ | $35 \; rad/s$ |
| Soft thetadot limit, $\dot{\theta}$ | $35 \; rad/s$ |
| Soft psidot limit, $\dot{\psi}$ | $35 \; rad/s$ |

*Table 1 - Dynamic System Parameters*

| Parameter | Default Value |
|---|---|
| Simulation Time Per Episode | 5 second |
| Simulation Frequency | 250 Hz |
| Control Frequency | 50 Hz |
| Initial State | [0, 0, 0, 0, 0, 0] |
| Random Seed (random seed if not set) | None |
| Constant Reference (random reference if not set) | None |
| Custom input limit (default input limits if not set) | None |
| Process Noise w mean and variance (only for stochastic environments) | (0, 0.01) |
| Measurement Noise v mean and variance (only for stochastic environments) | (0, 0.01) |
| Random seeds of the noises w and v (only for stochastic environment) | None |

*Table 2 - Environmental Parameters*

### 7) Done Condition

Time is the only designed end condition for this simulation. This is because the simulation states and inputs are bounded for given time, so there is no need for ending the simulation earlier than planned. Parameter $t_{start}$ and $t_{end}$ defines how long each episode will be.

Because of the implemented systems are time invariant, this time information only defines the episode length of the simulator.

### C. Configurable Parameters

The configurable parameters and their default values in the simulation framework are listed in Table 1 and Table 2:

## IV. DESIGN OF PERFORMANCE METRICS

For the performance metrics, the design should cover the classical control theory performance metrics, while it also gives meaningful comparison for the learning-based methods. Below, you can see the selected metrics.
- Computation Time
- Rise Time
- Settling Time
- Overshoot Percentage
- Peak Time
- Steady State Error
- Total Cost/Reward

## V. TRAINING FRAMEWORK, RESULTS AND COMPARISONS

To test the simulation framework, algorithm implementations from the Stable Baselines 3 [5] were used. Stable Baselines3(SB3) is a set of reliable and open-source implementations of reinforcement learning algorithms implemented by using the PyTorch [10] libraries. It's unified and tested structure makes it very easy to integrate and strong against potential bugs in the code. Below, the list of deep reinforcement learning algorithms used in testing and comparisons were shared:
- Deep Deterministic Policy Gradients (DDPG) [13]
- Proximal Policy Optimization (PPO) [14]
- Soft Actor Critic (SAC) [15]
- Twin Delayed DDPG (TD3) [16]
- Advantage Actor Critic (A2C) [17]

The learning procedure can be explained as follows. The main purpose of the learning framework is to allow agents to explore their state and action spaces effectively in order to optimize the next actions. For this aim, the learning framework should allow agents to explore many states efficiently. In Quad-Sim, the environment will randomly generate reference states in each episode and the agent will try to follow all reference signals to maximize its rewards. As a result, a generalized policy is generated to follow references for every state.

For all these algorithms, the agents were trained for 2 million time-steps to learn the reference tracking control objective. For the parameters of learning algorithms, the default parameters from the given references were used. For testing, step response control performance was measured in Figure 2. This plot was created using the QuadSim framework's default comparison function. The resulting performance metrics are also calculated and listed in Table 3.

| Performance Metric / Algorithm | PID | DDPG | PPO | SAC | TD3 | A2C |
|---|---|---|---|---|---|---|
| Computation Time (sec) | 2.804 | 4.310 | 4.230 | 4.531 | 3.990 | 4.176 |
| Rise Time (sec) | 0.420 | 0.400 | 0.220 | 0.160 | 0.240 | 0.400 |
| Settling Time (sec) | 3.280 | 0.640 | 0.960 | 1.160 | 1.220 | 1.980 |
| Overshoot Percentage (%) | 11.97 | 0 | 0.301 | 61.62 | 6.872 | 18.20 |
| Peak Time (sec) | 1.300 | 5 | 0.480 | 0.600 | 0.560 | 0.840 |
| Steady State Error (rad) | 0.003 | 0.328 | 0.052 | 0.256 | 0.054 | 0.236 |
| Total Reward (unitless) | -1.566 | -14.13 | -1.340 | -1.922 | -2.761 | -3.178 |

*Table 3 - Summary of performance metric results for trained RL algorithms. The parameters for the PID algorithm were tuned empirically and manually tuned, the performance of PID does not correspond to the optimal PID controller performance, it is used only for an example PID controller demonstration.*

## VI. CONCLUSION

In this study, a Python based quadcopter simulation framework has been designed and implemented. Then, this simulation framework has been tested with some existing state of the art deep reinforcement learning algorithms to show an example use-case. QuadSim simulation framework aims to be flexible and compatible with the existing machine learning and reinforcement learning tools. Since the main purpose of the resulting simulation framework was to study on the reinforcement learning based control algorithms, a fully customizable framework has been implemented without any code optimizations. Also, it is worth mentioning that the simulation framework code optimizations are not included in this research focus since the experiments showed that the GPU based training is more time-consuming than the simulation itself (when compared per time-step performance), therefore the simulation framework code optimizations are planned in future studies.

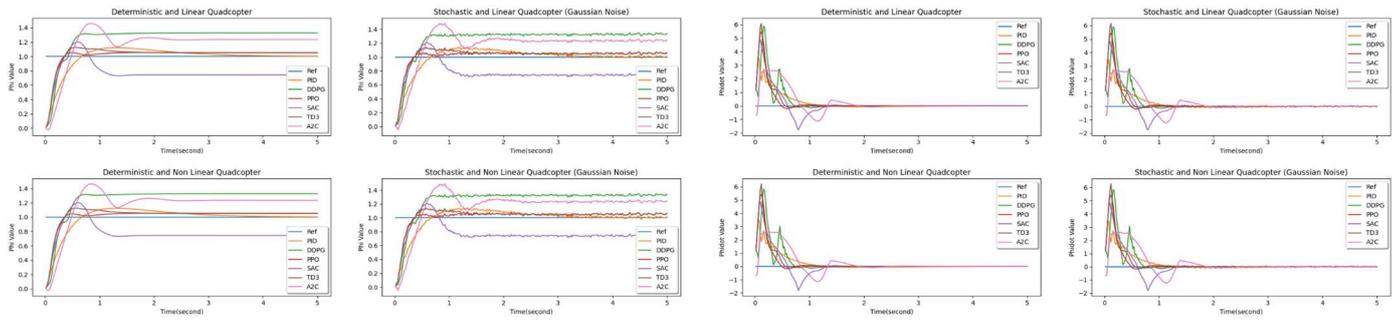

*Figure 2 - The roll angle(phi) and roll angular rate (phidot) step response results of PID controller and RL algorithms DDPG, PPO, SAC, TD3, A2C. RL algorithms were trained for 2M steps with default parameters. Each plot is labeled with their environment information and their corresponding state (phi in radians and phidot in radians/sec)*

As for the positive aspects of this work, after the simulation framework is downloaded, since the simulation framework is wrapped with "OpenAI Gym Environment", users will use this abstraction and they will be able to run their control algorithms without changing their own code. This will allow reaching many researchers and developers to use their own algorithms in the QuadSim simulation framework and compare their methods with others. Also, available algorithms are not limited with the reinforcement learning algorithms. Classical control and modern control theory algorithms can also be run in harmony. In addition, the simulation framework can be customized in terms of many features such as simulation dynamics, parameters, learning framework, etc.

The learning results were not examined in detail since the scope of this paper is limited with the simulation environment itself. While the resulting policy can follow the states generally, there is some steady state error in the roll state. This behavior can be fixed by increasing the Q matrix weight of the roll state while decreasing the weight of the roll angular rate state. This imperfect learning result is given by purpose to show the abilities and indicates the need for use simulation environments to improve learning performance iteratively. Further RL based controller performance and robustness will be investigated in future works.

In future studies, this framework design can be extended to have the feature of automatically executing experiments with many(like hundreds) of independent reinforcement learning training(which all of them cannot be executed in the same time because of the GPU VRAM usage of such trainings) and gathering the confidence intervals to robustly compare these algorithms with non-learning algorithms. Also, comparing the results with the most commonly used algorithms by the industry like PID, LQR, LQG, Linear MPC and Nonlinear MPC is necessary to show the performance metric results and see the real advantages/disadvantages of these RL based methods. The framework will provide such infrastructure for these studies, and it will open doors for further research.